\documentclass{svproc}
\usepackage{url}

\usepackage{algorithm}
\usepackage{algorithmicx}
\usepackage{algpseudocode}
\usepackage{graphicx}
\usepackage{hyperref}
\usepackage{amsfonts}
\usepackage{amsmath}
\begin{document}
\mainmatter 
\title{Depth Self-Optimized Learning Toward  \\Data Science}
\titlerunning{Depth Self-Optimized Learning Toward Data Science} 
%
\author{Ziqi Zhang\inst{1}}
\authorrunning{Ziqi Zhang} 
%
\tocauthor{Ziqi Zhang}
\institute{Tsinghua University, School of Life Science, China. \\
\email{zhangzq20@mails.tsinghua.edu.cn\\
stevezhangz@163.com}
}

\maketitle 

\begin{abstract}
We propose a two-stage model called Depth Self-Optimized Learning (DSOL), which aims to realize ANN depth self-configuration, self-optimization as well as ANN training without manual intervention. In the first stage of DSOL, it will configure ANN of specific depth according to a specific dataset. In the second stage, DSOL will continuously optimize ANN based on Reinforcement Learning (RL). Finally, the optimal depth is returned to the first stage of DSOL for training, so that DSOL can configure the appropriate ANN depth and perform more reasonable optimization when processing similar datasets again. In the experiment, we ran DSOL on the Iris and Boston housing datasets, and the results showed that DSOL performed well. We have uploaded the experiment records and code to our Github\footnote{\url{https://github.com/workharduwillwin/Depth-Self-Optimized-Learning-Toward-Data-Science} }.
\end{abstract}
\section{Introduction}
Date back to the 1989s, the universal approximation theorem states that a fixed depth neural network with arbitrary width and specific activation function such as sigmoid could approximate any continuous functions on a compact set to arbitrary accuracy\cite{gcb}\cite{kh:ms}. Approximate arbitrary functions might be the key point for a universal model, But when using this kind of ANN to process complex datasets, the number of nodes or layer width will reach exponential level, so this method is difficult to implement. Subsequently, people found that increasing the depth of neural network can make neural network easily approximate functions on some datasets that shallow neural network can't\cite{mt}\cite{re:os}\cite{hwl:mt:dr}\cite{tp:hm:lr}. Similar viewpoint could be found in Goodfellow's book$-$Deep learning\cite{gi:yb:ac} which states that in some cases, deeper networks can generalize better, and this is not just because of the larger number of parameters. Those views seems to be confirmed in real life, from AlexNet to RestNet, it is obvious that the performance of ANN is positively related to its depth, so it seems that only if we continuously increase the depth then we can gain a universal model.\\
Unfortunately, according to the No Free Lunch theorem (NFL)\cite{dh:wg}, no model can be fully qualified for various tasks, which seems to contradict our wishes. Even if we design a general model for a variety of different datasets within the allowable error range, we do not need to use more complex methods for those tasks can be completed by simple methods. Because it may waste too much computing resources. However, if we try to understand the NFL from another perspective, it seems that the theorem itself can be regarded as an excellent way to obviously improve the generalized ability of ANN, that is to say, using the best ANN according to specific dataset. \emph{In oreder to realize a universal mode based on the idea mentioned above, we mainly focus on solving those problems summarized as below:}
\begin{itemize}
\item[1]
How to design ANNs according according to specific datasets.
\item[2]
How to implement this design process through a model that can work independently.
\end{itemize}
Before this work, the above-mentioned second problem has been studied for a long time. In the past, people has realized Neural Architecture Search (NAS) based on RL, evolutionary algorithms, and so on \cite{te:jh}. Although the original intention of this type of research is to help people better search the hyperparameters of ANN, from another perspective, it is also an excellent way to realize ANN self-tuning. The first problem mentioned above may be more challenging, but we have gained some inspiration from human behavior, namely, we are trying to make ANNs understand what they will do. Based on these ideas and researches, we propose a two-stage model called Depth Self-Optimized Learning (DSOL). In the experiment, we used the Boston housing and Iris dataset to test the first stage of the DSOL. We set the True labels to 3, 10, 25, 50, 60, and train the first-stage for 100 iterations in turn. Experimental results show that the first stage of the DSOL can converge well on those two datasets, which indicates that the first-stage of DSOL has a excellent approximation and generalizion capability to some extent. Furthermore, when we normalize the training data, we find that the predictions of the model becomes more accurate. Then, we test the second stage of DSOL. Significantly, due to the limited performance of the device, we can only set the maximum number of ANN layers to 15, that is to say, if the prediction is greater than 15, then it will be regarded as 15. After training, the second stage of DSOL and ANN converged. In addition, by visualizing the number of ANN layers, it can be observed that well-trained DSOL tends to choose more deep ANNs to approximate on the dataset. \emph{Our contributions summarized as below:}
\begin{itemize}
\item[1]
We proposed a two-stage model called self-optimization learning (DSOL) for data science, which initially realizes ANN depth self-configuration and self-optimization. The advantage of this model is that it brings the datasets into the ANN parameter configuration process to know what they are doing. 
\end{itemize}
Although we have achieved staged victories, we still have a long way to go to achieve our ultimate goal. Our ultimate goal is to design a system that can automatically, accurately, and quickly generate, train, optimize ANNs for various specific datasets so as to realize a real universal model, namely, we only have to input dataset into it without doing any other things. In addition, this system can utilize various potential functions and modules, and can continuously upgrade the ANN in consideration of the performance of ANN and the feature of the dataset rather than only the depth of ANN. \emph{Now our model still has many limitations:}
\begin{itemize}
\item[1]
The second stage(RL-stage) of DSOL may fall into the cycle of local minimum in some cases. As the matter of fact, we have considered this problem, so we add a little randomness to the decision-making process of the second stage, and solve this problem to some extent, but it is not very efficient.
\item[2]
If some ANNs with the depth have an obvious different performance on the same dataset, the judgment of the RL model will be affected. However, the parameters of ANN should be initialized with an appropriate method to ensure it could attend the optimal level, namely, if we set all parameters to 0, the performance of ANN will be affected, which is contrary to our goal. So we must find a balance between the stability of RL and the performance of ANN.
\item[3]
The current architecture of the DSOL is not very complicated, so although it can be used on small-dimensional dataset, its performance on large-scale dataset is still unknown. At the same time, we know that the performance of ANN is not only related to its depth, so we need to further improve the model so that it can configure and optimize more kinds of ANN parameters.
\end{itemize}
In the introduction, we have briefly discussed the research background, solutions, experiments, contributions, and limitations of this research. The rest of this paper is as follows: In the second part of this paper, we will briefly introduce some researches related to this work. In the third part, we will introduce in detail the principles and architecture of DSOL. In the fourth part, we will analyze the experimental results. In the fifth part, we made a summary and stated our next research plan, and the last part is acknowledgement. Significantly, most of the experimental data and figures are in the attachment.
\section{Related work}
In the past five years, Reinforcement Learning (RL) has solved many problems that is difficult for traditional machine learning (ML). For example, RL has reached a superhuman level in Atari game \cite{mv:KK:etal} and poker game \cite{bn:ts}. In addition, RL has some practical applications, such as self-driving cars \cite{yy:xp:zw:cl}, and so on. In this paper, we mainly introduce its application in Neural Architecture Search (NAS).\\
Bbarret zoph et al. \cite {bz:qvl} used RCNN which is optimized by Reinforcement Learning (RL) based on policy gradient method to search for the best hyperparameters of ANN. They did it, but they used a lot of GPUs and ran for about a month, which was beyond the experimental conditions of ordinary people. Barret Zoph et al. \cite{zb:v:etal} proposed another method based on RL. They did not search for the complete ANN architecture, but first constructed a cell architecture, and then obtained an optimal ANN architecture constructed by these cell architectures. They used 500 GPUs and ran for about 4 days. These methods require a lot of computing resources, but there are some studies that have gradually reduced the requirement of computing source, such as some NAS researches based on Hierarchical Representation \cite{bz:etal}, Weight sharing\cite {ch:etal}, Performance prediction, and so on.\\
It is no doubt that NAS-related researches will make it more convenient for people to configure and optimize the hyperparameters of ANN. Our goal is not exactly the same as NAS, because we focus on developing a self-optimizing model rather than only help people configure hyperparameters of ANN. But we have some similarities, such as looking for a better ANN architecture.
\section{Methods}
\begin{figure}
\centering
\setlength{\abovecaptionskip}{0cm} 
\setlength{\belowcaptionskip}{-0.5cm}
\small
\includegraphics[scale=0.40]{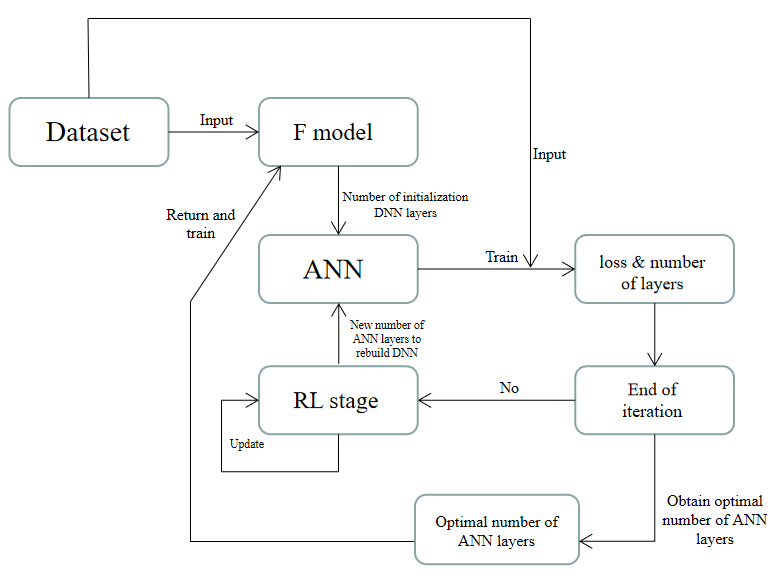}
\caption{The Self-Optimization Learning model.}
\label{fig:1}
\end{figure}
As shown in Fig.\ref{fig:1}, Algorithm \ref{alg}, DSOL includes two stages. In the first stage, the training dataset is input into the F-model to calculate the number of ANN layers, and then the RL model will optimize the number of layers according to the performance of the ANN. Significantly, ANN performance is realated to its hyperparameters, thus a deep ANN doesn't mean must be better than a shallow one. However, using some methods such as Batchnormalization or Dropout, and use Relu or LeakyRelu to replace sigmoid or no activation function will efficientlly solve the problem of gradient disappearance so as to improve the performence of ANN. In this paper, we use fixed 500 width Dense layer with Relu activation function, and every Dense layer followed by a Dropout layer to build ANN. Significantly, we initialize the bias of each Dense layer as 0, use Glorot Uniform methods to initialize the weights of ANN layers, and in the subsections, we will introduce the architecture and function of the DSOL in detail.
\begin{algorithm}[t]
\begin{algorithmic}
\State Initialize the state-policy function $ \pi $ weightens with Glorot Uniform method.
\State Initialize the policy-value function $\mathrm{Q}$ weightens with Glorot Uniform method.
\State Initialize replay buffer $h_t$ to capacity $N$.
\State Initialize the $F-model$ weights with Glorot Uniform method.
\State Initialize the minimum loss value $minloss$ to a large interger.
\For{episode $=1,M$} 
\State Initialize the number of ANN layers by $F\;model$, build a neural network $ANN_t$
\State Train ANN in fixed iterations $I$, return loss value $loss_t$.
\For {$t=1,T$}
\State With probability $\varepsilon$ select a random action $a_t$(Add or reduce a Dense layer).
\State Otherwise select $a_t\gets arg max(\pi(loss_t,layer_t;\theta))$.
\State Update $layer_t$ to $layer_{t+1}$ according to $a_t$.
\State Rebuild the neural network to $ANN_{t+1}$ according to $layer_{t+1}$.
\State Train $ANN_{t+1}$ in fixed iterations $I$, return loss value $loss_{t+1}$.
\If {$loss_{t+1} < minloss$} 
\State $minloss\gets loss_{t+1}$
\EndIf 

\State $$ r_t=\left\{
\begin{array}{ccc}
\frac{loss_t-loss_{t+1}}{loss_{t+1}}\times10+ \frac{minloss-loss_{t+1}}{minloss}\times10 $\qquad if$ & & {loss_{t+1}>loss_t}\\
\frac{loss_t-loss_{t+1}}{loss_{t}}\times10 $\qquad\qquad\qquad\qquad\qquad\qquad if$ & & {loss_{t+1}<loss_t}
\end{array} \right. $$
\State Store transition ($loss_t$, $layer_t$, $loss_{t+1}$, $layer_{t+1}$, $r_t$) in $h_t$.
\State Sample random minibatch {($loss_t$, $layer_t$, $loss_{t+1}$, $layer_{t+1}$, $r_t$)} from $h_t$.
\State Update the parameters of the actor based on equation 4.
\State Update the parameters of critics according to equation 5.
\EndFor
\State Return the optimal number of the ANN layers.
\State Update the parameters of the $F-model$ based on gradient descent.
\EndFor
\end{algorithmic}
\caption{Depth Self-Optimized Learning}
\label{alg}
\end{algorithm}
\subsection{F-model}
\begin{figure}
\centering
\setlength{\abovecaptionskip}{0cm} 
\setlength{\belowcaptionskip}{0cm}
\small
\includegraphics[scale=.43]{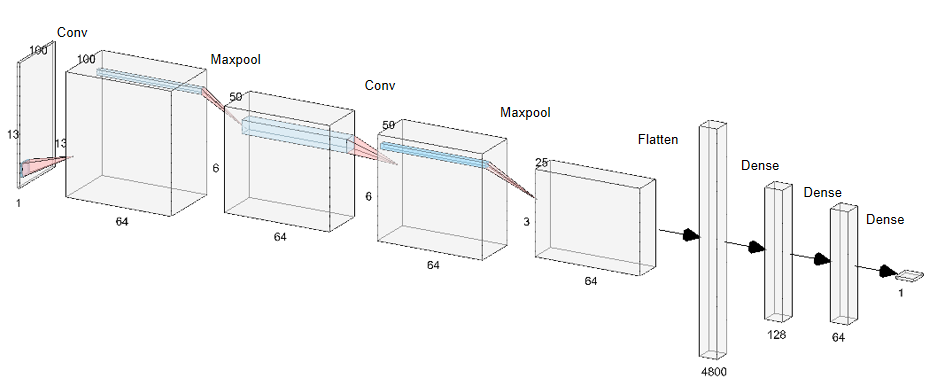}
\caption{ Architecture of the F-model. \emph{Using 100 samples from Boston dataset as training dataset, the architecture of F-model is shown in this figure.}}
\label{fig:2}
\end{figure}

The first stage of DSOL is F-model (Fig.\ref{fig:2}, Table.\ref{t1}, \ref{t2}). F-model consists of two convolutional layers, two pooling layers, and three or four fully-connected(FC) layers in turn. Significantly, according to the code we provided, there is a judgment in front of the FC layer that if necessary, add a Dense layer of appropriate width between the first FC and the last CNN layers so that the model can be successfully trained on various datasets(Table.\ref{t1}, Table.\ref{t2}). In addition, although the architecture of the F-model has been shown in this paper, we can use more powerful modules to replace it, such as combining RL with more powerful ANN to replace its current architecture. However, In order to successfully implement the ideas proposed in this paper, we will still use the initial design.
\subsection{Reinforcemnet Learning stage}
Now, we introduce the architecture of the Reinforcement Learning (RL) stage. The principle of the RL stage is similar to the agent-environment interaction model. At the beginning of the $t$ round of training, the environment will construct $ ANN_t $ according to the number of $ ANN_t $ layers $ layer_t $ which is configured by the F-model. After $ANN_t$ training, we obtain $loss_t$. In the experiment, both $loss_t $ and $layer_t $ are regarded as the state of the agent. Input the state $ loss_t $ and $ layer_t $ into the policy function $\pi$, and then we can get the next action $ a_t $(Equation \ref{e1}). Optimize $ ANN_t $ according to $ a_t $ so that obtain $ layer_ {t + 1} $(Equation \ref{e2}). Construct $ ANN_ {t + 1} $ according to $ layer_ {t + 1} $, after $ ANN_ {t + 1} $ training, we will get a new loss value $ loss_ {t + 1} $. If $loss_{t + 1}$ is greater than $loss_{t}$, the action will be punished, on the contrary, the action will be rewarded. Regardless of punishment or reward, all of these can be represented by $r_t$. Then, we will obtain the sequence data $h_t$ (Equation \ref{e3}), which will be used to train the policy function $\pi$ and the value function $Q $.
\begin{equation}
a_t=\pi(h_t;\theta)
\label{e1}
\end{equation}
\begin{equation}
\label{e2}
layer_{t+1}=layer_t+a_t
\end{equation}
\begin{equation}
\label{e3}
h_t=\{loss_1,layer_1,r_1,loss_2,layer_2,r_2,...,loss_t,a_t,r_t\}
\end{equation}
$\gamma$ is a constant decay rate range from 0 to 1(Here we use 0.8), and $r_{t}$ is a reward based on the performance of $ANN_t$. Therefore, the total reward in the RL training stage can be represented by the reward function: $ R_t = \sum_{t}^{\infty} \gamma ^ {{t'}-t} r_ {t'} $. In order to maximize $R_t$, we use the policy gradient (Equation \ref{e4}) to optimize the policy function $\pi$, and use the TD error (Equation \ref{e5}) to optimize the value function $Q$. Finally, we will obtain the optimal policy function $\pi^*$, and the value function $Q$ will also converge or reach the optimal level.
\begin{equation}
\label{e4}
\nabla_{\theta}\pi_{\theta}(s_t,a_t;\theta)=(r+\gamma Q_\pi(s_{t+1},a_{t+1};w)
-Q_\pi(s_t,a_t;w))
\nabla_{\theta}\log\pi_{\theta}(s_t,a_t;\theta)
\end{equation}
\begin{equation}
\label{e5}
loss(a_{t+1};W)=\frac{1}{2}\|r_t+\gamma Q_\pi(S_{t+1},a_{t+1};w)
-Q_\pi(S_t,a_t;w)\|^2
\end{equation}
\section{Experiment}
\subsection{F-model}
In order to test the approximation ability of the F model, we input the training and valid Boston housing dataset into the F-model, and set several integers: 3, 10, 25, 50, 60, 100 as True labels to train the F-model. It can be observed from Fig.\ref{fig:3} Fig.\ref{fig:4} that the F-model converges in all cases. Subsequently, input valid and training datasets into well-trained F-model in all cases for prediction. We can observe that the trained F-model has a good approximation ability(Fig.\ref{fig:4} and \ref{t5}). Therefore, if we can obtain the best number of ANN layers and regard them as True labels, then use the training dataset and True labels to train the  F-model, after training, when input similar dataset into the F-model, plausible predictions will be obtained. In addition, we normalized the training data to the range from 0 to 1, and trained the  F-model as before, we found that the approximation performance of the  F-model is better than before (Fig.\ref{fig:5}, Fig.\ref{fig:6} and Table.\ref{t6}). Although those experiments have proved that our model has good approximation ability, we pay more attention to the generalization ability of F model, because our original intention is to let ANN know what they want to do, so F model should be able to approximate on two or more than two datasets.\\
Take the Iris and Boston housing datasets as input in turn, and set the True labels to 3, 10, 25, 50, 60, 100 in turn, as shown in Fig.\ref{fig:7}, Fig.\ref{fig:8}, and Tbale.\ref{t7}. The F-model can converge on both these two datasets at the same time, which indicates that the F-model has a generalization ability. Significantly, the generalization ability of the F-model is of vital importance, because the feature of the training dataset is not directly related to the RL stage, so the generalization ability of DSOL is mainly reflected in the F-model.\\
\subsection{RL-stage}
First of all, let's restate the constraints of the ANN parameters. We set the initial layer number of ANN to 5 (Here we didn't use F-model to initialize the number of ANN layers because the F-model has not been trained. In the final test process of the experiment, we will use the trained F-model to initialize the layer number of ANN), and in the subsequent training process, the number of ANN layers should not exceed 15. In addition, we use the Golort Uniform method to initialize the weights of the ANN and set the bias of the initialized ANN layer to 0. In addition, except for the last and first layers of ANN, each Dense layer is followed by a Dropout layer with a probability of 0.5.\\
In the training process of the RL model, there are a total of 70 episodes, and in each episode, the RL model needs to optimize the ANN for 20 iterations. At the same time, we did not set the states of early termination for RL, because we have set the maximum number of ANN layers. According to related theories and some attempts, under the conditions we set, the performance of ANN increases with the number of layers. However, if we terminate the training of the model when the number of ANN layers reaches the maximum, we cannot prove that the RL model can stably set the ANN layer to the maximum in the subsequent training process, which means that the model may not have collected all states, so we should train the model as many times as possible without excessive interference. The experimental results are shown in Fig.\ref{fig:9}, Fig.\ref{fig:10}, and Fig.\ref{fig:11}. According to Fig.\ref{fig:10} and Fig.\ref{fig:11}, we can observe that the loss function and accuracy of the ANN gradually converge with the training process. According to Fig.\ref{fig:9}, Fig.\ref{fig:10}, and Fig.\ref{fig:11}, it can be observed that the performance of ANN is positively correlated with the increase in the number of ANN layers, and the well-trained RL model tends to add more Dense layers to the ANN, which is consistent with our expectations.\\
As mentioned in the methods section, we used the policy gradient ascent method to update the parameters of policy function $\pi$ and used TD error to optimize the parameters of the value function $Q$. Thus, both of them should converge after several episodes. Fortunately, the experimental results are consistent with the theory. As shown in Fig.\ref{fig:12}, we can observe that both $Q$ and $\pi$ converge into 0, which indicates that the RL stage has reached the optimal state. Significantly, maximizing the reward value of RL is equal to minimizing the reward value of RL multiplied by -1. Thus, in Fig. \ref{fig:12} the TD error converges to zero indicates the RL model has attended to the maximum reward level. Finally, we use the optimal number of ANN layers to train the F-model, and then get the trained DSOL. As shown in Fig.\ref{fig:13}, well trained DSOL performs well on the training dataset and can quickly optimize the number of layers of ANN to the optimal state, so that the loss function of ANN is always maintained at the lowest level.
\subsection{Validation}
We input 200 valid samples of the Boston housing dataset into a well-trained DSOL. As can be observed from Fig.\ref{fig:14}, at the beginning of ANN optimization, DSOL sets the number of ANN layers to 15, and then in the subsequent ANN training process, the number of ANN layers does not change. At the same time, the loss function and accuracy of the ANN don't fluctuate greatly. These results show that the neural network has reached the optimal level in the initial stage of training. Thus, the well-trained DSOL performs well on the valid Boston housing dataset.\\
\section{Conclusion and Perspective}
We propose DSOL for data science. The model initially realized ANN depth self-configuration and self-optimization, so that the best ANN can be obtained on a specific dataset. In the experiment, we use 100 Boston housing training dataset samples for training, and use 200 Boston housing valid dataset samples to test the trained DSOL. According to the discussion in the experimental section, DSOL performs well. Significantly, the generalization ability of DSOL is mainly reflected in the F-model. Therefore, an F-model with a certain generalization ability is equivalent to a DSOL with generalization ability. After testing the F-model, we found that the F-model can approximate well on the Iris and Boston housing datasets. Therefore, our DSOL has a certain generalization ability. \emph{However, DSOL is just an initial form. We hope to develop a self-optimized system to handle various datasets in the future. In order to achieve this, we will mainly focus on those things:} \\
\begin{itemize}
\item[1]
How to make the system optimize more kinds of ANN hyperparameters. We have known many NAS algorithms, which have been proved to be able to find new and feasible neural architectures. But we want to develop a model that can design the ANN architectures while taking datasets into account. That is to say, it's just like when a person processes a task, he will make a plan based on his analysis of the task.
\item[2]
How to make this system capable of dealing with more complex tasks rather than only data science. At present, most NAS algorithms can work on image datasets. However, this is a challenge for our model, because the image contains more complex information, so it may need to be improved with more complex logic to work.
\end{itemize}
\section{Acknowledgement}
We would like to acknowledge the two-month funding from the School of Lifescience in Tsinghua university, and thank one of my best friend Chuanxu Zhao for assistance in writing, and the reviewers for their valuable suggestions. Finally, best wishes to everyone who had ever encouraged and helpt me and everyone who are working hard for scientific research.
\newpage
\bibliographystyle{IEEEtran}

\begin{thebibliography}{}
\providecommand{\url}[1]{#1}
\csname url@samestyle\endcsname
\providecommand{\newblock}{\relax}
\providecommand{\bibinfo}[2]{#2}
\providecommand{\BIBentrySTDinterwordspacing}{\spaceskip=0pt\relax}
\providecommand{\BIBentryALTinterwordstretchfactor}{4}
\providecommand{\BIBentryALTinterwordspacing}{\spaceskip=\fontdimen2\font plus
\BIBentryALTinterwordstretchfactor\fontdimen3\font minus
  \fontdimen4\font\relax}
\providecommand{\BIBforeignlanguage}[2]{{%
\expandafter\ifx\csname l@#1\endcsname\relax
\typeout{** WARNING: IEEEtran.bst: No hyphenation pattern has been}%
\typeout{** loaded for the language `#1'. Using the pattern for}%
\typeout{** the default language instead.}%
\else
\language=\csname l@#1\endcsname
\fi
#2}}
\providecommand{\BIBdecl}{\relax}
\BIBdecl

\end{thebibliography}


\begin{thebibliography}{6}
\bibitem {gcb}
George Cybenko. Approximation by superpositions of a sigmoidal function. Mathematics of Control, Signals and Systems, 2(4):303–314.(1989)
\bibitem {kh:ms}
Kurt Hornik, Maxwell Stinchcombe, Halbert White, et al. Multilayer feedforward networks are universal approximators. Neural Networks, 2(5):359–366.(1989)
\bibitem {mt}
Matus Telgarsky. Benefits of depth in neural networks. In Conference on Learning Theory.(2016)
\bibitem {re:os}
Ronen Eldan,Ohad Shamir. The power of depth for feedforward neural networks. In Conferenceon Learning Theory.(2016)
\bibitem {hwl:mt:dr}
Henry W Lin, Max Tegmark,David Rolnick. Why does deep and cheap learning work so well?
Journal of Statistical Physics, 168(6):1223–1247.(2017)
\bibitem {tp:hm:lr}
Tomaso Poggio, Hrushikesh Mhaskar, Lorenzo Rosasco, Brando Miranda, and Qianli Liao. Why and
when can deep-but not shallow-networks avoid the curse of dimensionality: A review. International
Journal of Automation and Computing, 14(5):503–519.(2017)
\bibitem {gi:yb:ac}
Ian Goodfellow, Yoshua Bengio, Aaron Courville. Deep Learning. The MIT Press.(2016)
\bibitem {dh:wg}
D.H. Wolpert, W.G. Macready, No free lunch theorems for optimization. IEEE Transactions on Evolutionary Computation, vol. 1, no. 1, pp. 67-82.(1997) 
\url{doi: 10.1109/4235.585893.}
\bibitem {te:jh}
Thomas Elsken and Jan Hendrik Metzen and Frank Hutter. Neural Architecture Search: A Survey. arXiv preprint arXiv:1808.05377.(2019)
\bibitem{mv:KK:etal}
Mnih, V., K. Kavukcuoglu, D. Silver, A. A. Rusu, J. Veness, M. G.
Bellemare, A. Graves, M. Riedmiller, A. K. Fidjeland, G. Ostrovski,
et al.Human-level control through deep reinforcementlearning. Nature. 518(7540): 529–533.(2015)
\bibitem{bn:ts}
Brown, N. and T. Sandholm. Libratus: The Superhuman AIfor No-Limit Poker. International Joint Conference on Artificial Intelligence.(2017)
\bibitem{yy:xp:zw:cl}
You, Y., X. Pan, Z. Wang, and C. Lu.  
Virtual to Real Reinforcement Learning for Autonomous Driving. arXiv preprint arXiv:1704.03952.2017.(2017)
\bibitem {bz:qvl}
Barret Zoph, Quoc V. Le. Neural Architecture Search with Reinforcement Learning. arXiv preprint arXiv:1611.01578.(2016)
\bibitem{zb:v:etal}
Zoph, Barret and Vasudevan, Vijay and Shlens, Jonathon and Le, Quoc V. Learning Transferable Architectures for Scalable Image Recognition. Proceedings of the IEEE Conference on Computer Vision and Pattern Recognition (CVPR).(2018)
\bibitem{bz:etal}
Barret Zoph, Vijay Vasudevan, Jonathon Shlens, Quoc V. Le, Learning Transferable Architectures for Scalable Image Recognition. arXiv preprint arXiv:1707.07012.(2017)
\bibitem{ch:etal}
Cai H, Chen T, Zhang W, et al. Reinforcement Learning for Architecture Search by Network Transformation. arXiv preprint arXiv:1707.04873. (2017)
\end{thebibliography}

\appendix
\renewcommand{\appendixname}{Appendix~\Alph{section}}
\begin{table}
\section{Architecture of Depth Self-Optimizing Learning}
\caption{F-model on Boston housing dataset.}
\begin{center}
\begin{tabular}{ccc}
\hline\rule{0pt}{6pt}
Layer type & Shape& $\;\;\;\;\;\;\;\;\;$Param\\[2pt]
\hline \rule{0pt}{12pt}
Input& (1,100, 13, 1)&0\\
Conv2D& (1, 100, 13, 64)&1664 \\
Maxpooling&(1, 50, 6, 64)&0\\
Conv2D& (1, 50,6,64)&102464 \\
Maxpooling&(1, 25,3, 64)&0\\
Flatten&(4800)&0\\
Dense &(1,128)&614528\\
Dense &(1,64)&8256\\
Dense &(1,1)&65\\[2pt]
\hline
Total params: 726,977\\
Trainable params: 726,977\\
Non-trainable params: 0\\
\hline
\label{t1}
\end{tabular}
\end{center}
\caption{F-model trained on Iris dataset.}
\begin{center}
\begin{tabular}{ccc}
\hline\rule{0pt}{6pt}
Layer type & Shape&  $\;\;\;\;\;\;\;\;\;$Param\\[2pt]
\hline \rule{0pt}{12pt}
Input& (1,100, 13, 1)&0\\
Conv2D& (1, 100, 13, 64)&1664 \\
Maxpooling&(1, 50, 6, 64)&0\\
Conv2D& (1, 50,6,64)&102464 \\
Maxpooling&(1, 25,3, 64)&0\\
Flatten&(1600)&0\\
Dense(Relu) &(1,4800)&7684800\\
Dense(Relu) &(1,128)&614528\\
Dense(Relu) &(1,128)&614528\\
Dense(Relu) &(1,64)&8256\\
Dense(Relu) &(1,1)&65\\[2pt]
\hline
Total params: 8,411,777\\
Trainable params: 8,411,777\\
Non-trainable params: 0\\
\hline
\label{t2}
\end{tabular}
\end{center}
\begin{minipage}{\textwidth}
 \begin{minipage}[t]{0.45\textwidth}
  \centering
     \makeatletter\def\@captype{table}\makeatother\caption{Policy function $\pi$}
\label{t3}
       \begin{tabular}{ccc} 
    \hline \rule{0pt}{12pt}
Input& (1,2)&0\\
Dense(Relu) &(1,10)&614528\\
Dense(Relu) &(1,5)&8256\\
Dense(Softmax)&(1,2)&65\\[2pt]
\hline
Total params: 97\\
Trainable params: 97\\
Non-trainable params: 0\\
\hline
    \end{tabular}
  \end{minipage}
  \begin{minipage}[t]{0.45\textwidth}
   \centering
        \makeatletter\def\@captype{table}\makeatother\caption{Value function $Q$}
\label{t4}
         \begin{tabular}{ccc}        
          \hline \rule{0pt}{12pt}
Input& (1,2)&0\\
Dense(Relu) &(1,10)&614528\\
Dense(Relu) &(1,5)&8256\\
Dense &(1,2)&65\\[2pt]
\hline
Total params: 97\\
Trainable params: 97\\
Non-trainable params: 0\\
\hline

      \end{tabular}
   \end{minipage}
\end{minipage}
\end{table}
\begin{figure}
\section{Experiments of F-model}
\centering
\setlength{\abovecaptionskip}{0cm} 
\setlength{\belowcaptionskip}{0cm}
\small
\includegraphics[scale=.38]{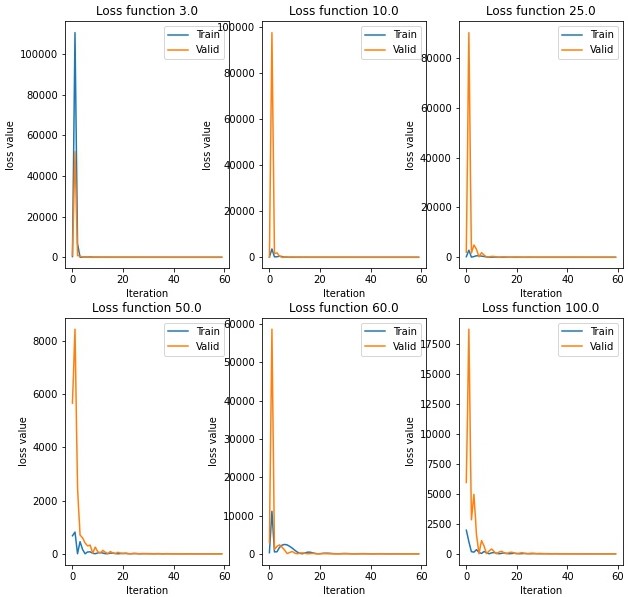}
\caption{Visualization of F-model loss function on Boston housing dataset. \emph{The Boston hosing dataset is used as the input data, the true labels are set to integers: 3, 10, 25, 50, 60, 100 so as to establish the mapping from the input data to those integers by training F-model in turn. Then the loss function is plot in turn. From these functions, we could observe that the F-model converges in all cases.}}
\label{fig:3}
\centering
\includegraphics[scale=.38]{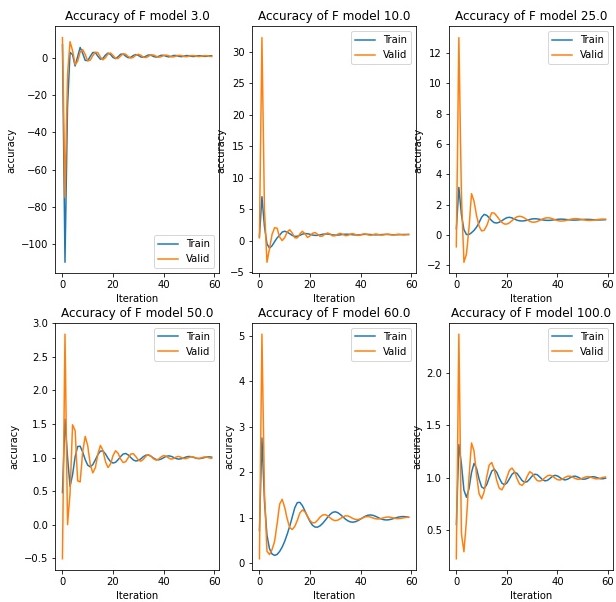}
\caption{Visualization of F-model accuracy on Boston housing dataset. \emph{Input the Boston housing valid dataset into the trained F-model, we obtain the predictions, then by dividing the predicted value with the true value, we get the precision shown in all cases.(True label as 3, 10, 25, 50, 60, 100 in turn)}}
\label{fig:4}
\end{figure}
\begin{figure}
\centering
\setlength{\abovecaptionskip}{0.25cm} 
\setlength{\belowcaptionskip}{0.25cm}
\small
\includegraphics[scale=.38]{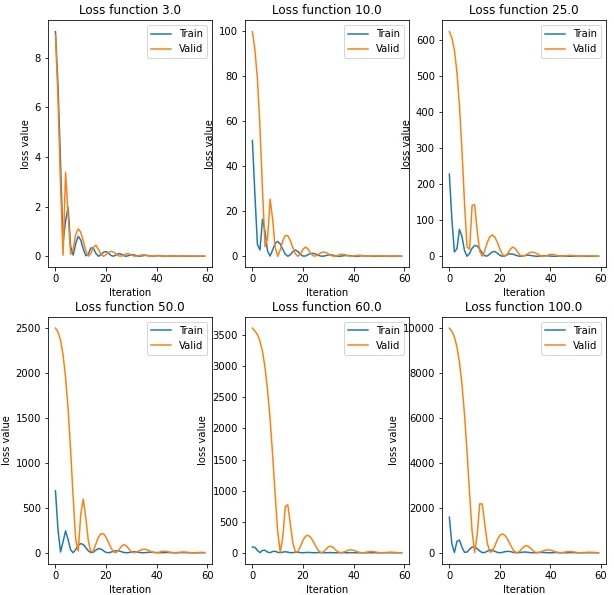}
\caption{Visualization of F-model loss function on normalized Boston housing dataset. \emph{After normalizing the Boston housing dataset, we observe an obvious enhancement of the approximation capability of F-model.}}
\label{fig:5}
\centering
\includegraphics[scale=.38]{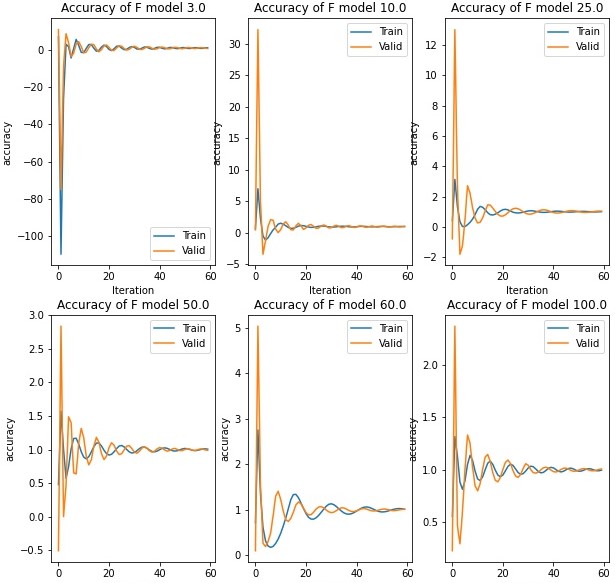}
\caption{Visualization of F-model accuracy on normalized Boston housing dataset. \emph{Input the normalized Boston housing valid dataset into the trained F-model, we obtain the predictions, then by dividing the predicted value with the true value, we get the precision shown in all cases.(True label as 3, 10, 25, 50, 60, 100 in turn)}}
\label{fig:6}
\end{figure}
\begin{figure}
\centering
\setlength{\abovecaptionskip}{0.25cm} 
\setlength{\belowcaptionskip}{0.25cm}
\small
\includegraphics[scale=.38]{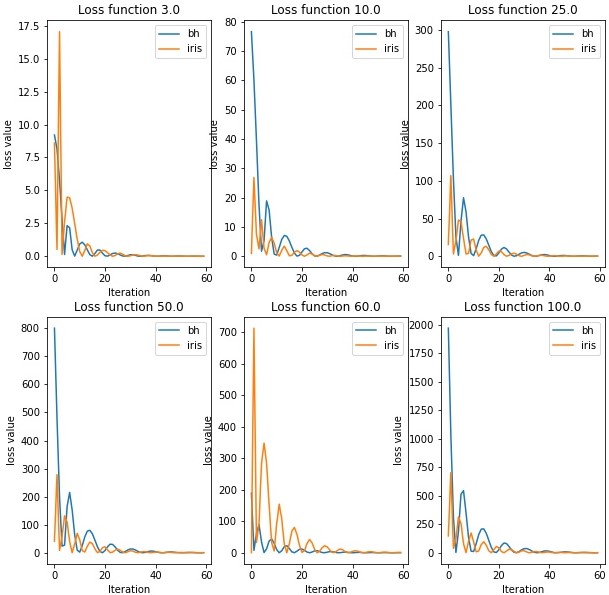}
\caption{Visualization of F-model loss function on normalized Iris and Boston housing dataset. \emph{In order to test the generalization capability of F-model, we use the Iris and Boston housing datasets as the Input in turn, and we can oberserve that F-model converge on both of Iris dataset.}}
\label{fig:7}
\centering
\includegraphics[scale=.38]{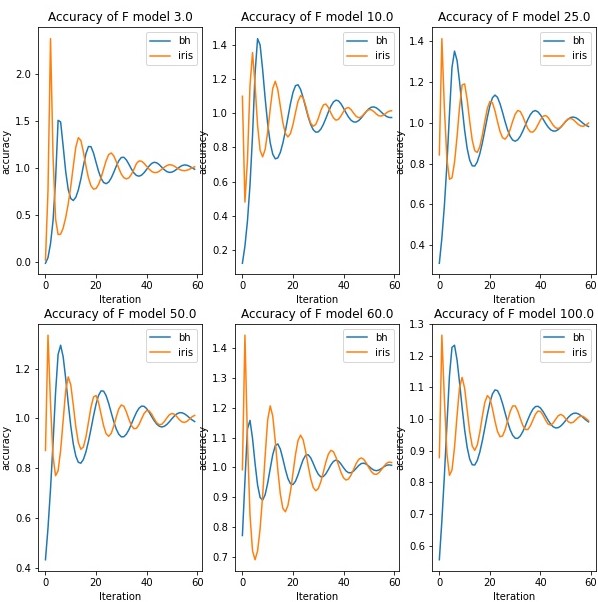}
\caption{Visualization of F-model accuracy on normalized Iris and Boston housing datasets. \emph{Using Iris and Boston housing dataset to train F-model in turn, we can observe that F-model approximate well on Iris and Boston housing datasets.}}
\label{fig:8}
\end{figure}
\begin{table}
\caption{Performance of F-model on Boston housing dataset \emph{Take the Boston housing training dataset as input, and take 3, 10, 25, 50, 60, 100 as real labels to train F-model in turn. The prediction results are shown in this table. In the first column, these predictions are obtained by inputting the training dataset into the trained F-model. In the second column, these predictions are obtained by inputting the valid dataset into the trained F-model. In the last column, these values are the true labels configured by ourselves. }}
\label{t5}
\begin{center}
\begin{tabular}{ccc}
\hline\rule{0pt}{6pt}
Train dataset$\;$& Valid dataset$\;$& True layer\\[2pt]
\hline \rule{0pt}{12pt}
2.897& 5.592& 3.0\\
9.886& 10.145 &10.0 \\
25.218& 22.807 &25\\
49.928&50.005 &50.0\\
60.159 &58.391 &60.0 \\
100.466 & 104.244 &100.0 \\[2pt]
\hline
\end{tabular}
\end{center}
\caption{Performance of F-model on normalized Boston housing dataset. \emph {We normalize the Boston housing dataset to the range from 0 to 1, and still set integers: 3, 10, 25, 50, 60, 100 as true labels in turn. After training, we found that the approximation ability of F-model has been significantly improved. So we should normalize the training data before inputting it into F-model.}}
\label{t6}
\begin{center}
\begin{tabular}{ccc}
\hline\rule{0pt}{6pt}
Train dataset$\;$& Valid dataset$\;$& True layer\\[2pt]
\hline \rule{0pt}{12pt}
2.944& 2.936& 3.0\\
9.932& 10.139 &10.0 \\
24.962&25.426 &25\\
49.766&50.902 &50.0\\
59.958 &60.906 &60.0 \\
98.728 &100.162 &100.0 \\[2pt]
\hline
\end{tabular}
\end{center}
\caption{Generalization capability of F-model. \emph {In order to test the generalization ability of the F-model, we use Iris and Boston housing training datasets as input to train F-model in turn and then use iris and Boston housing test training datasets as input to obtain the predictions shown in the table. The first column is the prediction obtained by inputting Boston housing valid dataset into the F-model, the second column is the prediction obtained by inputting the Iris test dataset into the F-model, and the third column is the True label we set.}}
\begin{center}
\begin{tabular}{ccc}
\hline\rule{0pt}{6pt}
Boston housing$\;$& Iris$\;$& True layer\\[2pt]
\hline \rule{0pt}{12pt}
1.301& 3.140& 3.0\\
8.225& 10.300 &10.0 \\
23.073& 25.889 &25\\
49.090&51.984 &50.0\\
59.011 &62.214 &60.0 \\
97.988 &101.951 &100.0 \\[2pt]
\hline
\label{t7}
\end{tabular}
\end{center}
\end{table}
\begin{figure}
\section{Experiments of RL-stage}
\centering
\setlength{\abovecaptionskip}{0cm} 
\setlength{\belowcaptionskip}{0cm}
\small
\includegraphics[scale=.38]{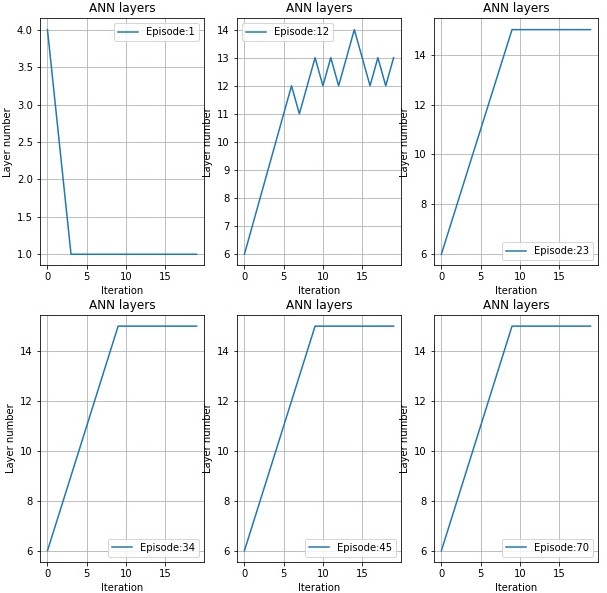}
\caption{Visualization of ANN layers. \emph{During the RL training process, we set up 70 episodes, and each episode contains 20 iterations. By plotting the changes in the number of ANN layers from episodes 1, 12, 23, 34, 45, and 70, we can intuitively observe that as the number of sets increases, the RL model tends to add more layers to the ANN.}}
\label{fig:9}
\centering
\includegraphics[scale=.38]{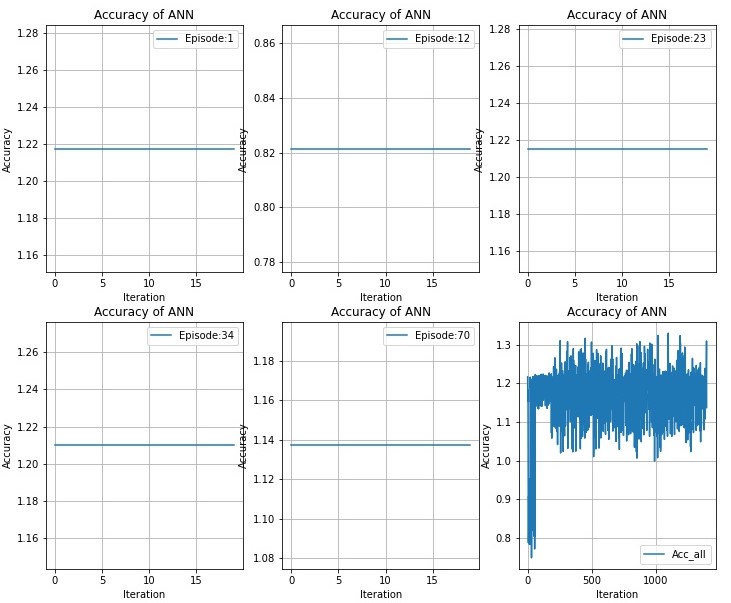}
\caption{Visualization of the accuracy of ANN. \emph{We sample sequential accuracy values from 1, 12, 23, 34, and 70 episodes in turn, and then plot these samples and total accuracy. In the last picture, we can roughly get the change trend of accuracy. During the ANN training process, its accuracy value gradually increases to about 1, and there is no abrupt change after that. It is worth noting that in order to obtain the accuracy value, we use the predicted value divided by the actual value, and then take the average value to represent the accuracy. This is only an approximate representation method, so the value obtained by this method may be greater than 1.}}
\label{fig:10}
\end{figure}
\begin{figure}
\centering
\includegraphics[scale=.32]{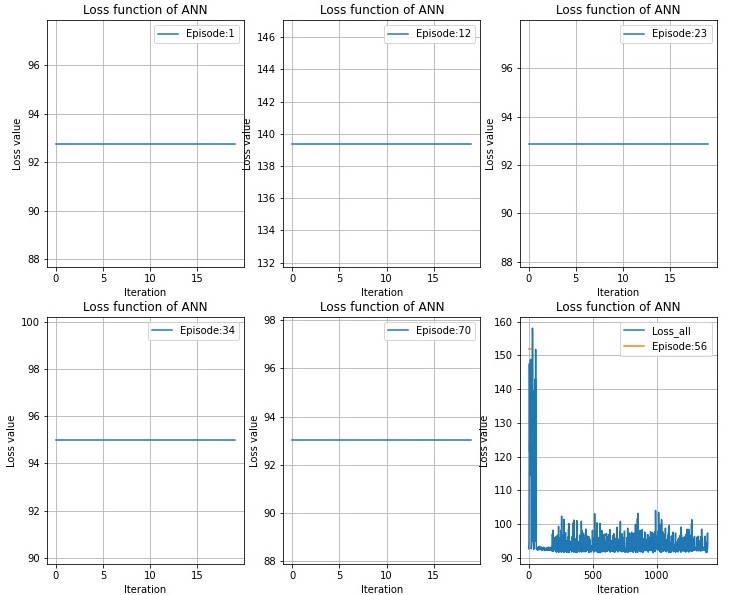}
\caption{Visualization of the loss function of ANN. \emph{The loss function of ANNs correspond to Fig.10.}}
\label{fig:11}
\centering
\setlength{\abovecaptionskip}{0.1cm} 
\setlength{\belowcaptionskip}{0cm}
\small
\includegraphics[scale=.38]{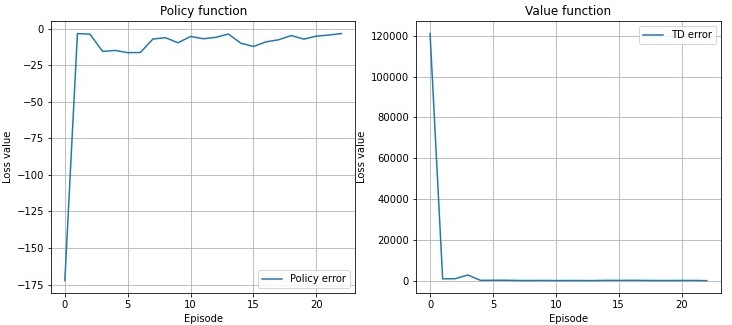}
\caption{\emph{As shown in the left figure the loss value of the Policy function $\pi$ converges to 0, which proves that $\pi$ has attend to the most stable state $\pi^*$. Significantly, during the experimental process, in order to maximize the TD value, we multiply TD by - 1 and reduce the processed TD value by the Actor Critic method. Therefore, as shown in the right figure when the processed TD value converges to 0, it also indicates that the benefit of the model reaches the maximum at this time.}}
\label{fig:12}

\centering
\includegraphics[scale=.38]{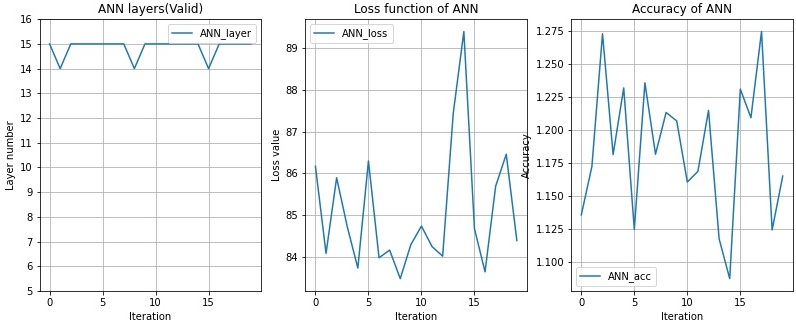}
\caption{Visualization of trained DSOL.\emph{Using trained DSOL to configure and optimize the ANN. We can observe that the number of ANN layers is optimized to the optimal level quickly by trained DSOL.}}
\label{fig:13}
\end {figure}
\newpage
\begin{figure}[t]
\section{Validation of trained DSOL}
\centering
\setlength{\abovecaptionskip}{0.1cm} 
\setlength{\belowcaptionskip}{0cm}
\includegraphics[scale=.38]{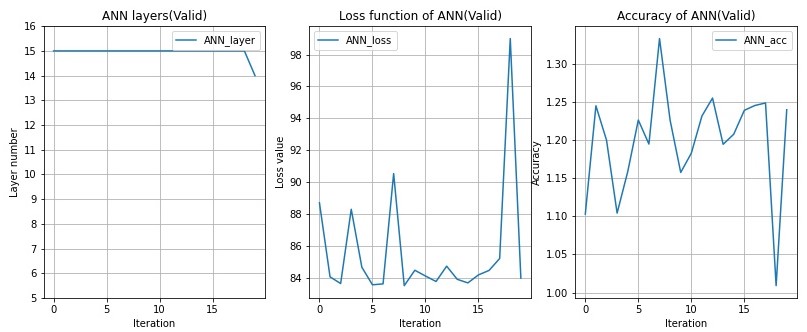}
\caption{Validation of trained DSOL \emph{Input 200 samples of valid Boston Housing dataset into the trained DSOL, record the performance of ANN and DSOL. We can observe that DSOL performs well on the valid Boston Housing dataset.}}
\label{fig:14}
\end{figure}
%
%

%

\end{document}